\documentclass[10pt,twocolumn,letterpaper]{article}

\PassOptionsToPackage{hyphens}{url}

\usepackage{graphicx}
\usepackage{amsmath}
\usepackage{amssymb}
\usepackage{booktabs}
\usepackage{float}
\usepackage{algorithm}
\usepackage{algpseudocode}
\usepackage{subcaption}
\usepackage{stfloats}
\usepackage[pagebackref,breaklinks,colorlinks]{hyperref}
\usepackage[capitalize]{cleveref}
\crefname{section}{Sec.}{Secs.}
\Crefname{section}{Section}{Sections}
\Crefname{table}{Table}{Tables}
\crefname{table}{Tab.}{Tabs.}

\usepackage{wacv}              % To produce the CAMERA-READY version

\begin{document}

%%%%%%%%% PAPER ID  - PLEASE UPDATE
\def\wacvPaperID{00007} % *** Enter the WACV Paper ID here
\def\confName{WACV}
\def\confYear{2025}

\title{Advancing Super-Resolution in Neural Radiance Fields via Variational Diffusion Strategies}
\author{Shrey Vishen\\
Monta Vista High School\\
Cupertino, California\\
{\tt\small shrey.vishen@gmail.com}
\and
Jatin Sarabu\\
Bellarmine College Preparatory\\
Santa Clara, California\\
{\tt\small jatinsarabu13@gmail.com}
\and
Saurav Kumar\\
UI Urbana-Champaign\\
Urbana, Illinois\\
{\tt\small sauravk4@illinois.edu}
\and
Chinmay Bharathulwar\\
John P Stevens High School\\
Edison, New Jersey\\
{\tt\small cbharathulwar3@gmail.com}
\and
Rithwick Lakshmanan\\
Pleasant Valley High School\\
Bettendorf, Iowa\\
{\tt\small rithwick.laks@gmail.com}
\and
Vishnu Srinivas\\
Foothill High School\\
Pleasanton, California\\
{\tt\small vishnuvishaks@gmail.com}
}

\maketitle
\begin{abstract}
   We present a novel method for diffusion-guided frameworks for view-consistent super-resolution (SR) in neural rendering. Our approach leverages existing 2D SR models in conjunction with advanced techniques such as Variational Score Distilling (VSD) and a LoRA fine-tuning helper, with spatial training to significantly boost the quality and consistency of upscaled 2D images compared to the previous methods in the literature, such as Renoised Score Distillation (RSD) proposed in DiSR-NeRF \cite{lee2024_disrnerf}, or SDS proposed in DreamFusion. The VSD score facilitates precise fine-tuning of SR models, resulting in high-quality, view-consistent images. To address the common challenge of inconsistencies among independent SR 2D images, we integrate Iterative 3D Synchronization (I3DS) from the DiSR-NeRF framework \cite{horstmeyer2021differentiable}. Quantitative benchmarks and qualitative results on the LLFF dataset demonstrate the superior performance of our system compared to existing methods such as DiSR-NeRF. All our code is available at \url{https://github.com/shreyvish5678/SR-NeRF-with-Variational-Diffusion-Strategies}
\end{abstract}

\section{Introduction}
\label{sec:intro}

Neural Radiance Fields (NeRFs) have revolutionized 3D scene rendering from 2D images, significantly impacting applications such as 3D reconstruction and virtual reality \cite{mildenhall2020nerf}. NeRFs use continuous volumetric functions optimized by neural networks to synthesize high-fidelity views \cite{sun2022direct}\cite{yu2021pixelnerf}\cite{tewari2022advances}. However, scaling NeRF for super-resolution, maintaining view consistency, and managing high-dimensional data remains challenging \cite{yu2021plenoxels}\cite{zhang2020nerf++}\cite{tancik2022block}\cite{xiangli2021citynerf}. Somewhat recent advancements such as Mip-NeRF 360 \cite{barron2022mip} and TensoRF have addressed some of these challenges with NeRF, especially with unbounded scene rendering and tensor decomposition. Our work focuses specifically on enhancing super-resolution capabilities within the NeRF framework. 

Previous enhancements to NeRF, like Score Distillation Sampling (SDS), have struggled with issues such as over-smoothing and computational inefficiency, limiting their ability to capture fine details. To address these problems, we introduce Variational Score Distillation (VSD), which models 3D scene parameters as probabilistic distributions rather than fixed values. This approach improves scene representation by leveraging diffusion models \cite{ho2020denoising}\cite{nichol2021improved} and incorporates low-rank adaptation (LoRA) for efficient fine-tuning of pre-trained models.

Our extensive experiments show that VSD significantly outperforms SDS and RSD in generating detailed and photorealistic NeRFs, enhancing visual quality and computational efficiency. Results from datasets like LLFF confirm these improvements. Additionally, we provide a detailed ablation study on our system's components, including LoRA-based fine-tuning \cite{hu2022lora} and hierarchical sampling strategies\cite{shreyvish2024advancing}.

Our approach advances NeRF-based rendering, offering a new standard for high-resolution 3D scene generation with broad applications in entertainment, gaming, scientific visualization, and architectural design \cite{barron2021mip}\cite{martin2021nerf}\cite{pumarola2021d}.

To summarize, the contributions of this paper are as follows,
\begin{itemize}
    \item Using Pre-trained Stable Diffusion weights for 3D NeRF Render Upscaling
    \item Utilizing Low-rank adaptation with Mixed Precision training for steering outputs
    \item Using a version of VSD loss and I3DS for training and upscaling the NeRF outputs
\end{itemize}

\section{Methodolgy}

\subsection{Pre-Requisites}

\subsubsection{Latent Encoding and Residual Learning}
Latent encoding starts by extracting 2D projections from a lower-resolution Neural Radiance Field (NeRF). These 2D views are then passed through an encoder, transforming them into latent space representations that capture key features in a compressed form \cite{rombach2022high}.

To improve image quality, we introduce learnable residual latents—additional vectors added to the original encoding. During training, these residuals are adjusted to correct errors or enhance specific features, refining the latent vectors over time. This iterative process, where residuals update with the model, boosts both image quality and consistency. The combined latents look like this:

\begin{equation}
  x_{0}' = x_{0} + h_{\theta}
\end{equation}

\subsubsection{Forward Diffusion Process}
Once the latent vectors, including the residual latents, are generated, they undergo a forward diffusion process \cite{song2020score}. Noise is gradually added to the latent vectors over timesteps, governed by the equation:

\begin{equation}
  x_t = \sqrt{\alpha_t} x_0 + \sqrt{1 - \alpha_t} \epsilon, \quad \epsilon \sim \mathcal{N}(0, I)
\end{equation}

This transforms the structured latent representation into a noisy state by the final timestep. The diffusion process simulates the challenge of reversing noise to recover the original image details, essential for the model to learn the data distribution.

In the subsequent denoising step, the noisy latents are refined to produce high-quality, view-consistent images \cite{ho2020denoising}\cite{nichol2021improved}. The overall goal of this process is to prepare the latent vectors for reconstruction, reversing the noise added during forward diffusion.

\subsubsection{UNet-Based Prediction}

In the model, the prediction of the final high-quality image relies heavily on the use of UNet architectures. The UNet models are employed to process the noisy latent vectors generated during the forward diffusion process and to reconstruct these into cleaner, more accurate representations of the original scene.

The first stage of latent prediction uses a frozen, pre-trained UNet model trained on a large image denoising dataset. This makes it well-suited for processing the noisy latent vectors generated by the forward diffusion process. The pre-trained UNet takes as input the noisy latent vector \( \mathbf{x}_t \), time embeddings \( \mathbf{t} \), text embeddings \( \mathbf{y} \) with applied class labels \( \mathbf{c} \), and the low-resolution image \( \mathbf{I}_{LR} \), aiming to predict the noise added to the latent vector \({\mathbf{x}}_0 \) also called \(\epsilon\). 

Mathematically, this is expressed as:
\begin{equation}
  \epsilon_{\phi} = f_{\phi}(\mathbf{x}_t, \mathbf{t}, \mathbf{y}^\mathbf{c}, \mathbf{I}_{LR})
\end{equation}

where \( f_{\phi} \) is the function representing the pre-trained model, and \(\phi\) represents its parameters.

This output serves as a reference for evaluating the fine-tuned model's performance. 

The noisy latent vector \( \mathbf{x}_t \) is also processed by a Fine-Tuned UNet model, which includes learnable Low-rank Adaptation (LoRA) parameters \cite{hu2022lora}. These allow the model to efficiently adapt for high-quality image reconstruction. Unlike the frozen pre-trained model, the Fine-Tuned UNet is trained during this stage, with LoRA parameters enabling targeted adjustments to the network layers.

The Fine-Tuned UNet takes the same inputs: \( \mathbf{x}_t \), time embeddings \( \mathbf{t} \), text embeddings \( \mathbf{y} \), and low-resolution image \( \mathbf{I}_{LR} \), plus class labels \( \mathbf{c} \) to focus on task-specific features \cite{lugmayr2021srflow}. Its prediction is:
\begin{equation}
  \epsilon_{\varphi} = f_{\varphi}(\mathbf{x}_t, \mathbf{t}, \mathbf{y}, \mathbf{I}_{LR}, \mathbf{c})
\end{equation}

where \( f_{\varphi} \) represents the function learned by the Fine-Tuned UNet with LoRA and \(\varphi\) represents its parameters.

To quantify the improvement in image quality provided by the Fine-Tuned UNet, we compute the Variational Score Distillation (VSD) loss, which measures the difference between the predictions of the pre-trained and fine-tuned models. We used an L1 variation of the VSD loss proposed in Prolific Dreamer \cite{wang2023prolificdreamer}, defined as:
\begin{equation}
  \mathcal{L}_{\text{VSD}}(\theta) = \mathbb{E}_{t, \epsilon, \mathbf{c}} \left[ \omega(t) \cdot \left\| \epsilon_{\phi} - \epsilon_{\varphi} \right\| \right]
\end{equation}
where \( \omega(t) \) is a weighting function that adjusts the importance of the loss based on the timestep \( t \). 

This loss is then backpropagated through the network, specifically targeting the LoRA parameters in the Fine-Tuned UNet. By minimizing this loss, the model learns to generate high-quality latent representations that closely match the ideal outputs while incorporating task-specific adjustments through the class labels. Then we obtrain the new residual latents:
\begin{equation}
  \theta_{\text{new}} = \theta_{\text{old}} - \eta \nabla_{\theta} \mathcal{L}_{\text{VSD}}(\theta)
\end{equation}

To add, every few steps, or every step, the LoRA parameters are fine-tuned as well, with the following equation:
\begin{equation}
  \mathcal{L}_{\text{Diff}}(\theta) = \mathbb{E}_{t, \epsilon, \mathbf{c}} \left[ \left( f_{\varphi}(\mathbf{x}_t, t, \mathbf{c}, \mathbf{I}_{LR}, \mathbf{y}) - \epsilon \right)^2 \right]
\end{equation}

Where, \(\mathbf{x}_t\) was obtained using the forward diffusion process with \(\mathbf{x}_0\) and \(\epsilon\). Then we can backpropagate this loss into the LoRA parameters, likewise:
\begin{equation}
  \varphi_{\text{new}} = \varphi_{\text{old}} - \eta \nabla_{\varphi} \mathcal{L}_{\text{Diff}}(\theta)
\end{equation}

Now the refined latent vector \( \hat   {\mathbf{x}}_0^{\text{'}} \) from the Fine-Tuned UNet and the Pre-trained one is iteratively improved through the training process, leading to progressively higher-quality image outputs. The final prediction is not only a denoised version of the latent vector but also one that has been optimized for the specific rendering task at hand, thanks to the targeted adjustments made possible by the LoRA parameters. 

This dual UNet approach, leveraging both pre-trained and fine-tuned models, ensures that the latent prediction process is both accurate and adaptable, ultimately contributing to the generation of high-resolution, view-consistent images that surpass the quality of those produced by existing methods \cite{lugmayr2021srflow}\cite{wang2018esrgan}.

\subsection{Low-rank Adaptation (LoRA) Fine-Tuning}

Low-rank Adaptation (LoRA) is a powerful technique designed to fine-tune pre-trained neural networks efficiently, particularly in scenarios where extensive retraining or structural modifications are impractical \cite{hu2022lora}. In the context of Neural Radiance Fields (NeRFs), LoRA introduces trainable low-rank matrices into selected layers of the network, allowing for effective adaptation to new datasets or specific rendering tasks with minimal computational overhead.

The core idea behind LoRA is to augment the existing layers of a pre-trained NeRF model with additional trainable parameters that capture essential modifications without altering the original weights. Specifically, LoRA integrates two low-rank matrices, denoted as \( A \in \mathbb{R}^{m \times r} \) and \( B \in \mathbb{R}^{r \times n} \), where \( r \) is a rank that is significantly smaller than the dimensions of the original weight matrix \( W \in \mathbb{R}^{m \times n} \). The low-rank matrices \( A \) and \( B \) are introduced in such a way that the original weight matrix \( W \) is modified as follows:
\begin{equation}
  W' = W + AB
\end{equation}

Here, \( W' \) represents the new effective weight matrix after the application of LoRA. This adjustment allows the model to learn additional features or adapt to new data without having to retrain the entire network from scratch.

The low-rank matrices \( A \) and \( B \) are fine-tuned during the training process, while the original weights \( W \) remain fixed. This approach ensures that the adaptation process is both efficient and effective, targeting only the parameters necessary for the specific task. The gradients for the matrices \( A \) and \( B \) are computed using standard backpropagation techniques, with the loss function \( \mathcal{L} \) defined for the specific rendering task, such as image enhancement, 3D reconstruction, or scene understanding. The gradients are given by:
\begin{equation}
  \frac{\partial \mathcal{L}}{\partial A} = \frac{\partial \mathcal{L}}{\partial W'} B^T, \quad \frac{\partial \mathcal{L}}{\partial B} = A^T \frac{\partial \mathcal{L}}{\partial W'}
\end{equation}

These gradients guide the updates to \( A \) and \( B \) during the training process, allowing the NeRF model to fine-tune its performance on the new task without modifying the extensive pre-trained weights \( W \) .

\begin{figure}
    \centering
    \includegraphics[width=1\linewidth]{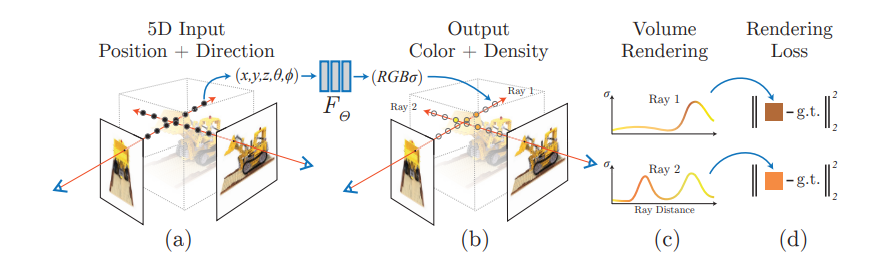}
    \caption{NeRF diagram}
    \label{fig:enter-label}
\end{figure}

\subsection{Lower Resolution NeRF Generation}
The first step in our pipeline involves constructing a low-resolution (LR) Neural Radiance Field (NeRF) from an LR image dataset. Using Nvidia's InstantNGP framework \cite{mueller2022instant}, we efficiently build this NeRF model. Afterward, the LR images are upsampled to match the resolution of the target super-resolution (SR) images, establishing a baseline for the iterative training process \cite{park2021nerfies}\cite{reiser2021kilonerf}\cite{chen2022tensorf}.

\subsection{SR Training Process}
The training process builds upon recent NeRF advancements, utilizing insights from a variety of volumetric rendering techniques \cite{wang2021neus}\cite{yariv2021volume} and neural field applications \cite{xie2022neural}. The training loop operates iteratively until convergence, following a multi-step approach:

1. \textbf{Random Render Sampling}: A random LR image is rendered from the NeRF and used as input for encoding. This image is then taken and interpolated to 4x it's size, and then converted into a latent. Both the image and latent are passed. A text prompt is passed as well which describes the NeRF content along with from what orientation the image should look like, to condition the model for the desired output.

2. \textbf{Learnable Residual Latents}: Trainable residual latents are added to the latent encoding, for helping to refine  image quality over time.

3. \textbf{Forward Diffusion}: The combined latents are passed through the forward diffusion process, to get the noisy latents, along with a given timestep embedding.

4. \textbf{Pre-trained UNet Prediction}: The noisy latent, along with text and time embeddings, is passed through a frozen, pre-trained UNet, which predicts a latent output used as a reference. The noisy latent is also processed by a fine-tuned UNet model with Low-Rank Adaptation (LoRA) parameters to adapt the network efficiently. Class labels are added to guide the learning process.

5. \textbf{VSD Loss Calculation}: The Variational Score Distillation (VSD) loss, based on the L1 loss between the pre-trained and fine-tuned UNet outputs, drives the optimization of the residual latents, by backpropagating this loss to the residual latents.

6. \textbf{Noise Differentiation Loss to LoRA}: Every few iterations, an auxiliary noise differentiation loss further refines the LoRA parameters. LoRA parameters are updated through backpropagation with this loss to improve task-specific fine-tuning.

This process repeats until the model converges, producing high-resolution, view-consistent NeRF outputs that outperform existing methods \cite{mittal2012making}\cite{wang2021ibrnet}, as we can see in Algorithm 1

\begin{algorithm}
\caption{VSD Super Resolution}
\label{alg:vsd_super_resolution}
\begin{algorithmic}[1]
\State \textbf{Inputs:} Latent $x_0$, text prompt embeddings $y$, timestep $t$, LR Image $I_{lr}$, class labels $c$, max timesteps $M$
\State \textbf{Outputs:} Latent residuals $h_{\theta}$
\State Initialize $h_{\theta}$
\For{$S = [0, M]$}
    \State $\epsilon \sim \mathcal{N}(0, 1)$
    \State $x_{0}' = x_0 + h_{\theta}$
    \State $x_t = \sqrt{a_t}x_{0}' + \sqrt{1 - a_t}\epsilon$
    \State $\epsilon_{\phi} = f_{\phi}(x_t, t, y^c, I_{lr})$
    \State $\epsilon_{\varphi} = f_{\varphi}(x_t, t, y, I_{lr}, c)$
    \State $\mathcal{L}_{VSD} = \mathbb{E}_{t, \epsilon, c}\left[w(t) \cdot \|\epsilon_{\phi} - \epsilon_{\varphi}\|\right]$
    \State $\theta \leftarrow \theta - \eta_{1} \nabla_{\theta} \mathcal{L}_{VSD}$
    \State $\mathcal{L}_{Diff} = \mathbb{E}_{t, \epsilon, c}\left[\left(f_{\varphi}(x_t, t, y, I_{lr}, c) - \epsilon\right)^2\right]$
    \State $\varphi \leftarrow \varphi - \eta_{2} \nabla_{\varphi} \mathcal{L}_{Diff}$
\EndFor
\end{algorithmic}
\end{algorithm}

\begin{figure}[H]
    \centering
    \includegraphics[width=1\linewidth]{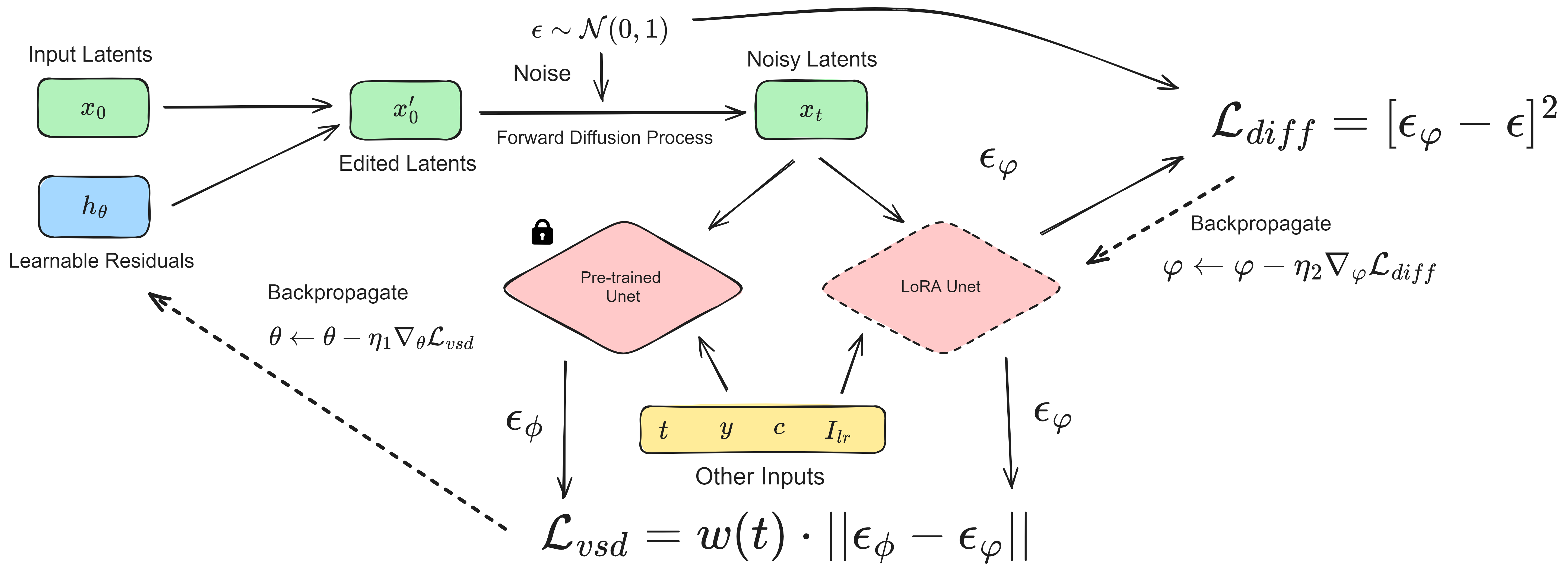}
    \caption{Refer to Algorithm 1}
    \label{alg1}
\end{figure}

\subsection{Iterative 3D Sync Process}

The Iterative 3D Synchronization (I3DS) approach addresses limitations observed in initial experiments, where applying SDS directly on NeRF renders resulted in blurred details and convergence issues. We add I3DS to the pipeline to improve this by decoupling the upscaling and NeRF synchronization processes into two alternating stages.

\begin{itemize}
    \item \textbf{Upscaling Stage:} Starting with a low-resolution NeRF, images are rendered at 4x resolution. These images are then independently upscaled using RSD to add high-resolution details. However, initial upscaling may produce inconsistent details across views.
    
    \item \textbf{Synchronization Stage:} The upscaled images are used as inputs to update the NeRF model. During this stage, NeRF is trained using standard procedures to synchronize view-consistent details, correcting inconsistencies introduced during upscaling.
\end{itemize}

The synergy between upscaling and synchronization stages allows for increasingly detailed and view-consistent outputs over iterations. I3DS efficiently balances these stages to optimize for high-quality, consistent NeRF outputs while minimizing memory requirements and improving convergence times. This method demonstrates a reduction in optimization duration by 4x compared to previous approaches. We can see further about I3DS in Algorithm 2.

\begin{algorithm}
\caption{Iterative 3D Synchronization (I3DS)}
\label{alg:i3ds}
\begin{algorithmic}[1]
\State \textbf{Input:} LR NeRF $\omega_{l}$, LR images $I_{lr}$, training poses $P_{tr}$
\State \textbf{Output:} SR NeRF $\omega_{sr}$

\State $\omega \gets \omega_{l}$
\For{$\text{S} = [0, M]$}
    \State \textbf{Upscaling Stage}
    \State $x_0 \gets \text{RenderImage}(\omega, P_{tr})$
    \State $x_0 \gets \text{InterpolateX4}(x_0)$
    \State $z_0 \gets \text{VaeEncode}(x_0)$
    \State $z_0' \gets \text{VSD}(z_0, I_{lr})$
    \State $x_0' \gets \text{VaeDecode}(z_0')$
    \State $I_{tr} \gets x_0'$
    
    \State \textbf{Synchronization Stage}
    \For{$\text{sync\_iter} = [0, \text{max\_sync\_iter}]$}
        \State $(r_o, r_d, c_{tr}) \gets \text{SampleRays}(I_{tr}, P_{tr})$
        \State $c' \gets \text{RenderRays}(r_o, r_d)$
        \State Perform gradient descent on $\nabla_{\omega} \| c' - c_{tr} \|$
        \State $\omega_{old} \gets \omega$
    \EndFor
\EndFor

\State \textbf{Return} $\omega_{sr} \gets \omega$
\end{algorithmic}
\end{algorithm}

\begin{figure}[H]
    \centering
    \includegraphics[width=1.0\linewidth]{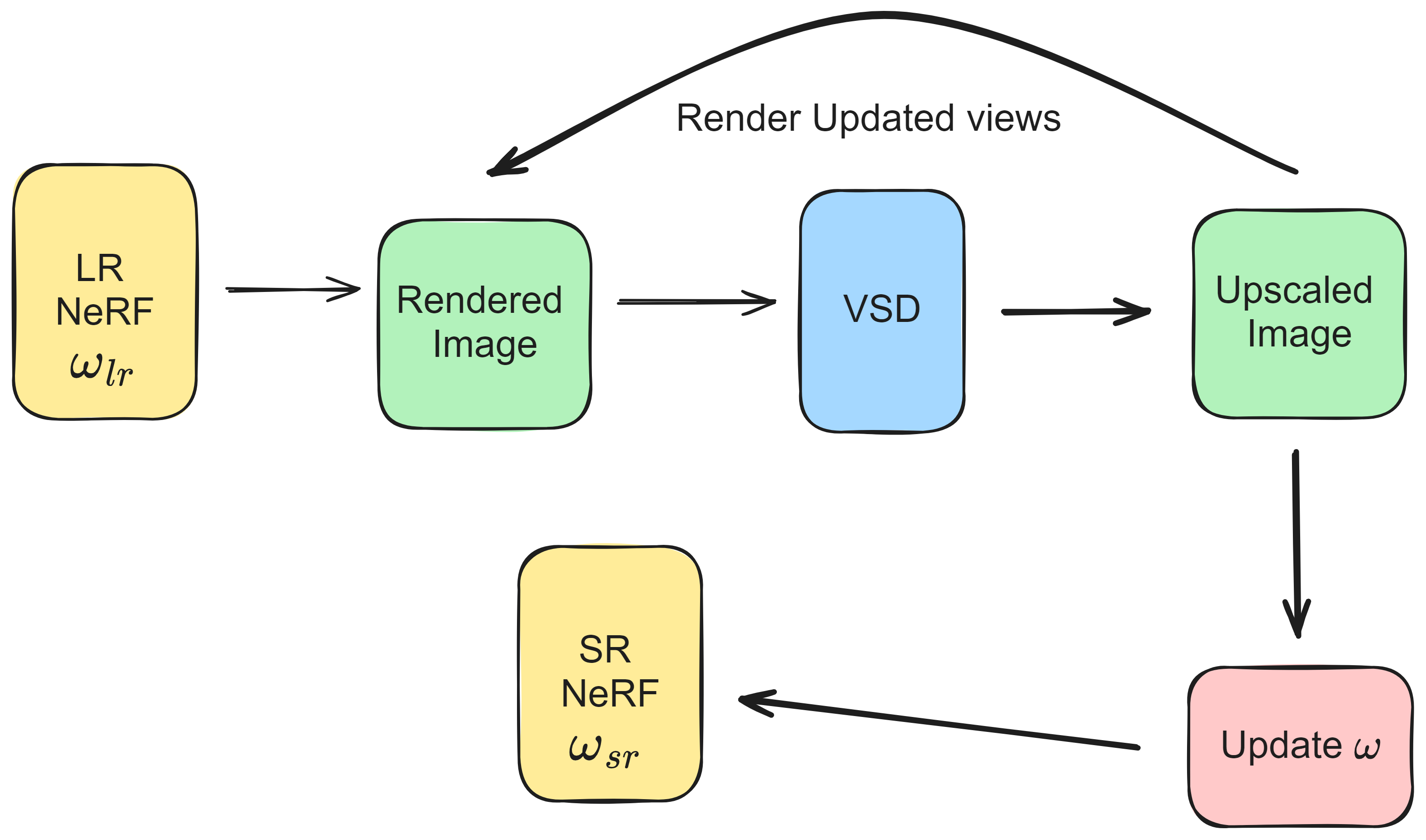}
    \caption{Refer to Algorithm 2}
    \label{alg2}
\end{figure}

\section{Experimental Results}

Images of a museum artifact and a flower were generated using RSD, SDS, and VSD. VSD was used to generate images of both the artifact and the flower, whereas RSD was used exclusively for the artifact, and SDS was used solely for the flower. Across all 3 methods, there are noticeable differences in color saturation, texture, and detail.

\begin{figure*}[t]
    \centering
    \begin{subfigure}[b]{0.24\textwidth}
        \includegraphics[width=\linewidth]{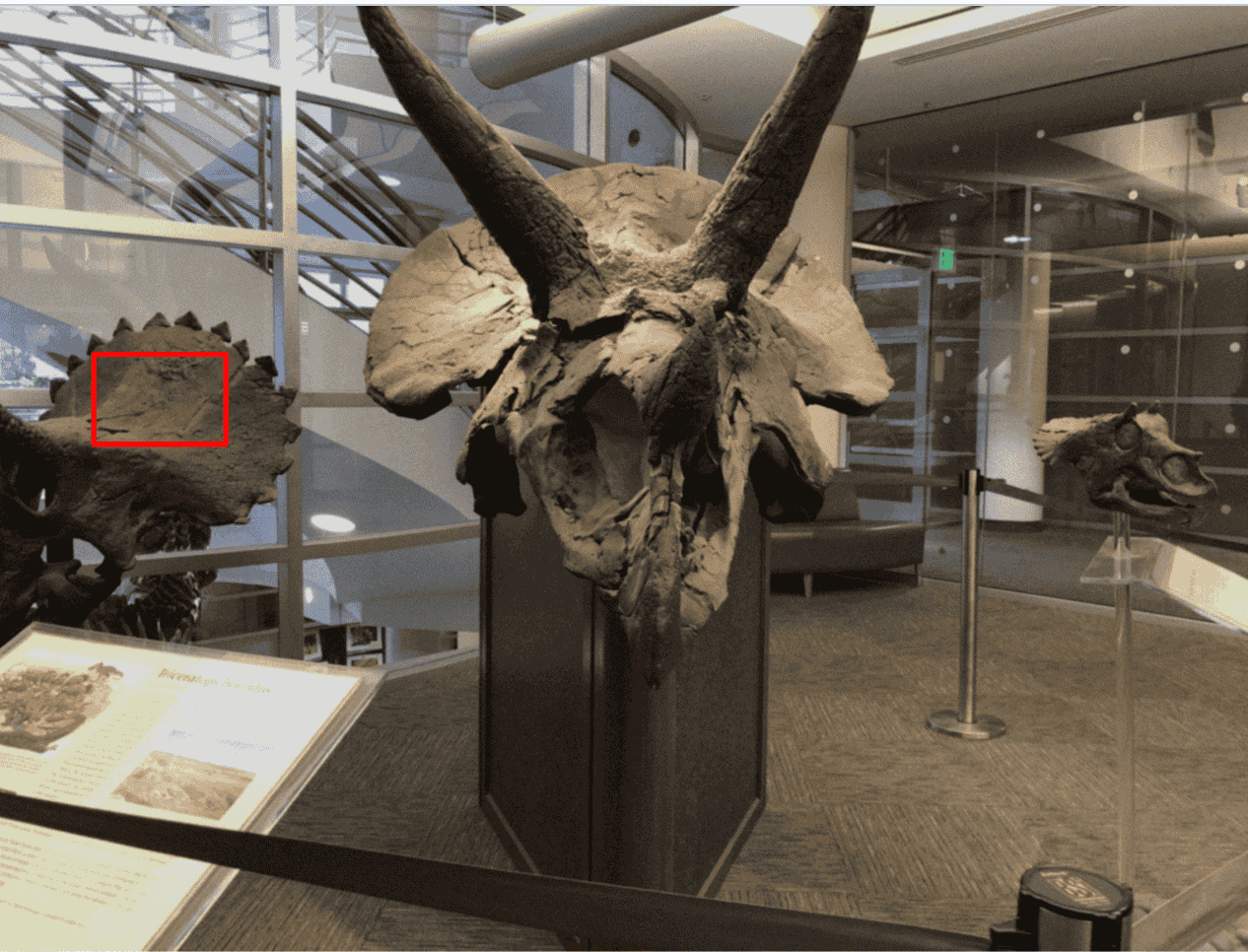}
        \caption{RSD Artifact Results}
    \end{subfigure}
    \begin{subfigure}[b]{0.24\textwidth}
        \includegraphics[width=\linewidth]{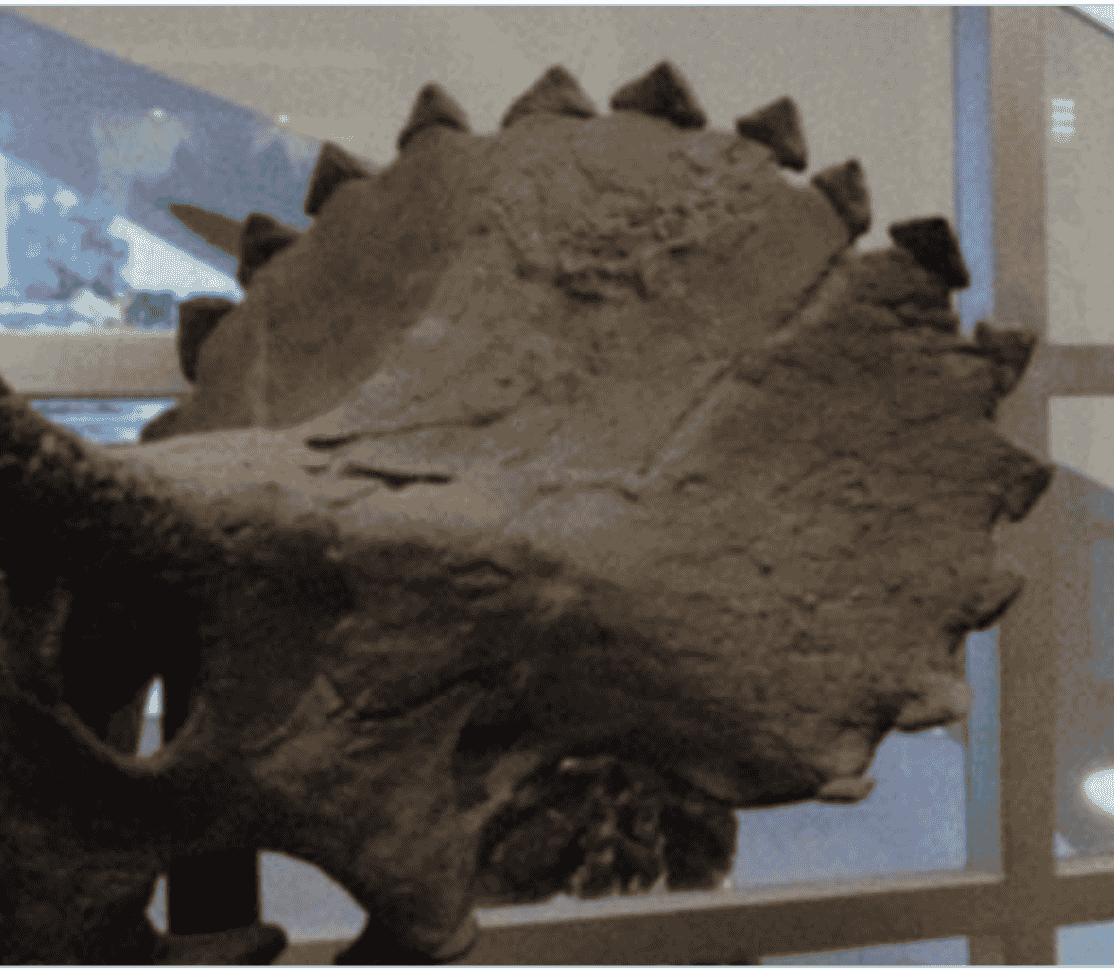}
        \caption{RSD Artifact (Zoomed)}
    \end{subfigure}
    \begin{subfigure}[b]{0.24\textwidth}
        \includegraphics[width=\linewidth]{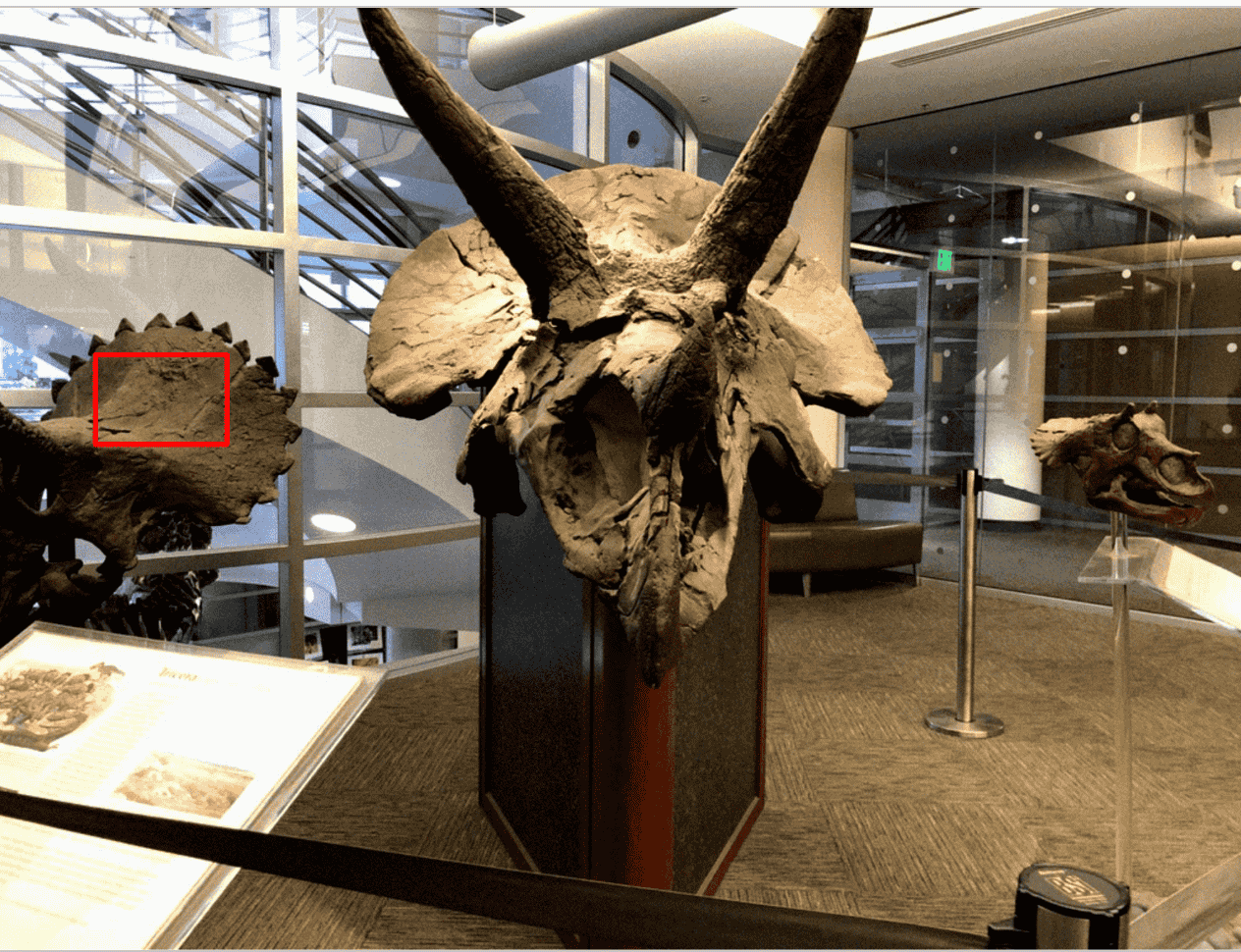}
        \caption{VSD Artifact Results}
    \end{subfigure}
    \begin{subfigure}[b]{0.24\textwidth}
        \includegraphics[width=\linewidth]{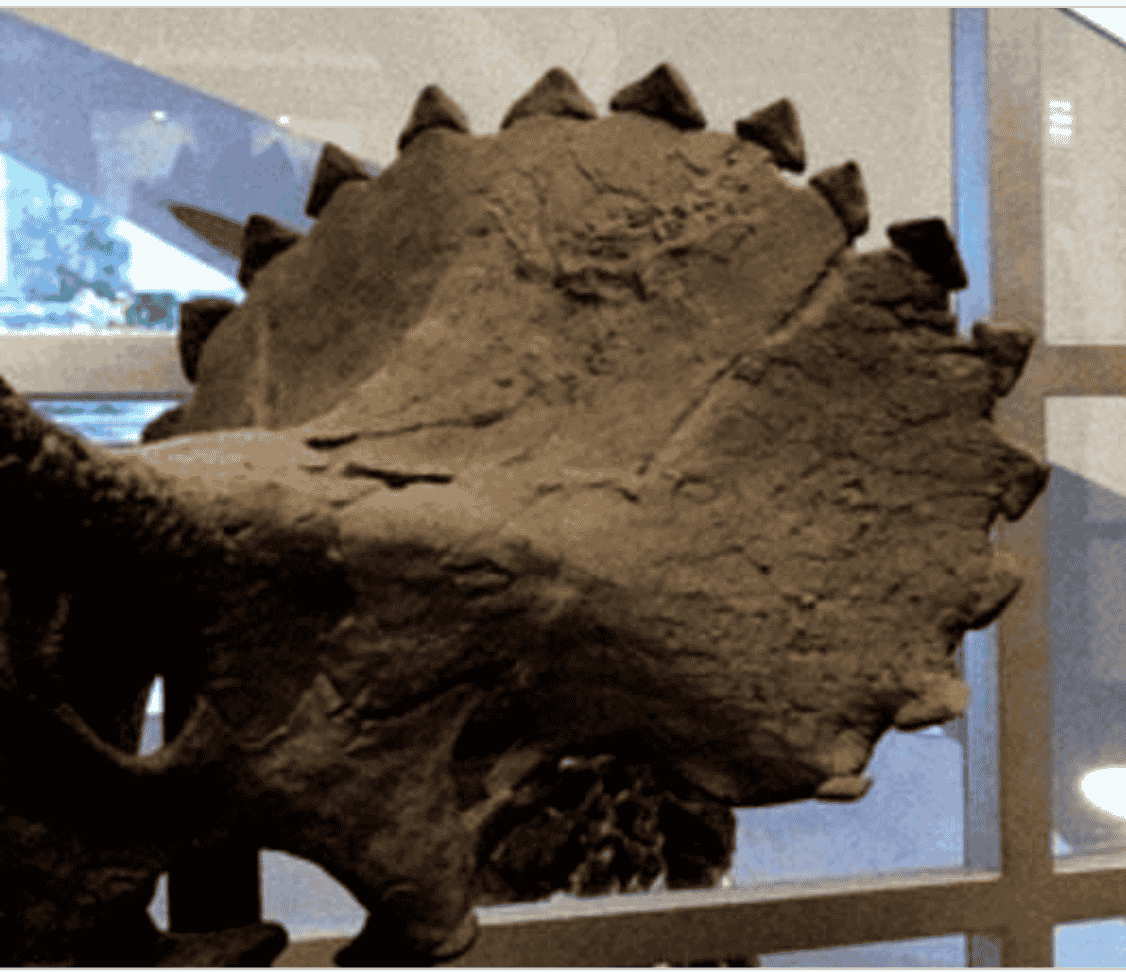}
        \caption{VSD Artifact (Zoomed)}
    \end{subfigure}
    \caption{Comparison of RSD and VSD on Museum Artifact}
    \label{fig:artifact-comparison}
\end{figure*}

\begin{figure*}[t]
    \centering
    \begin{subfigure}[b]{0.24\textwidth}
        \includegraphics[width=\linewidth]{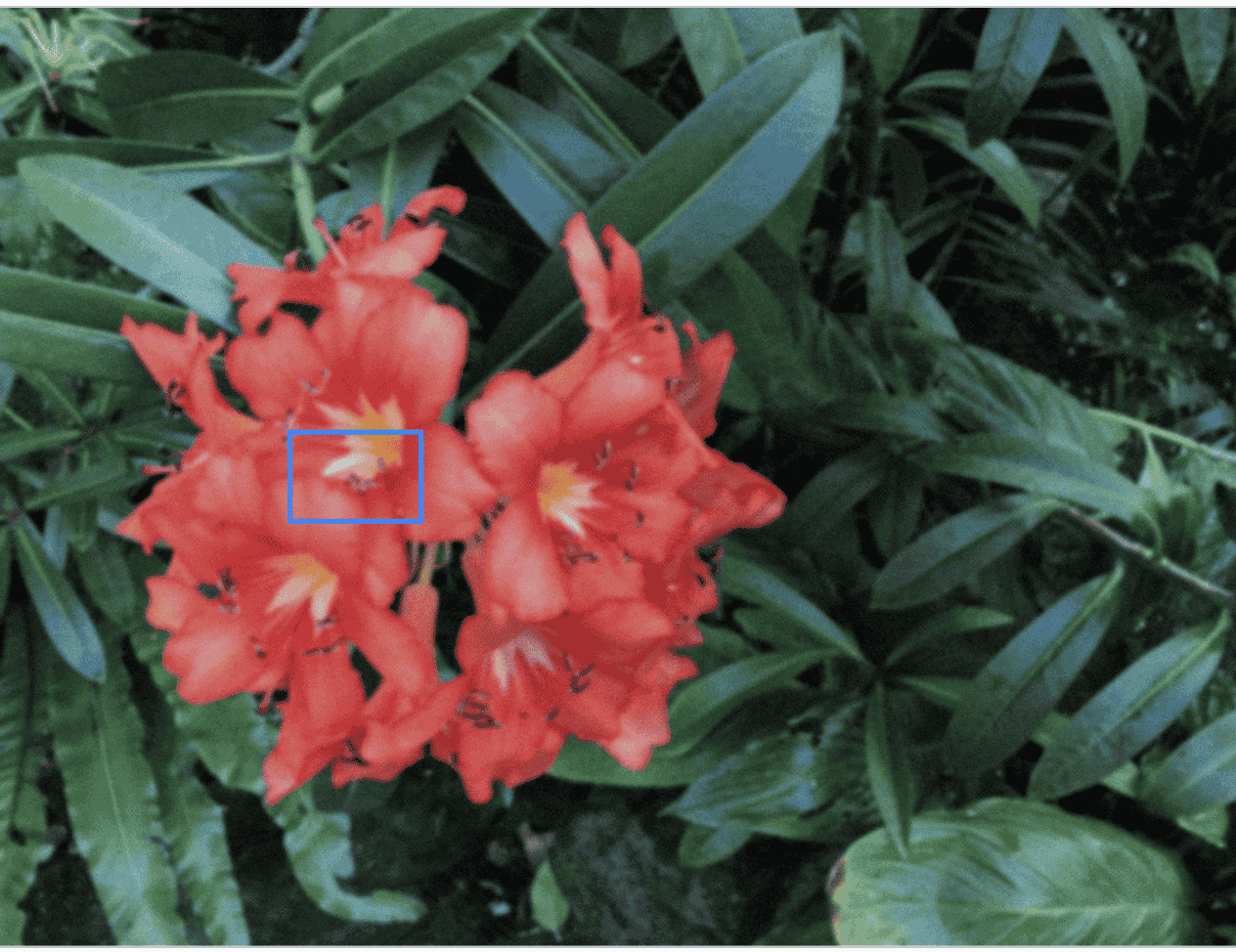}
        \caption{SDS Flower Results}
    \end{subfigure}
    \begin{subfigure}[b]{0.24\textwidth}
        \includegraphics[width=\linewidth]{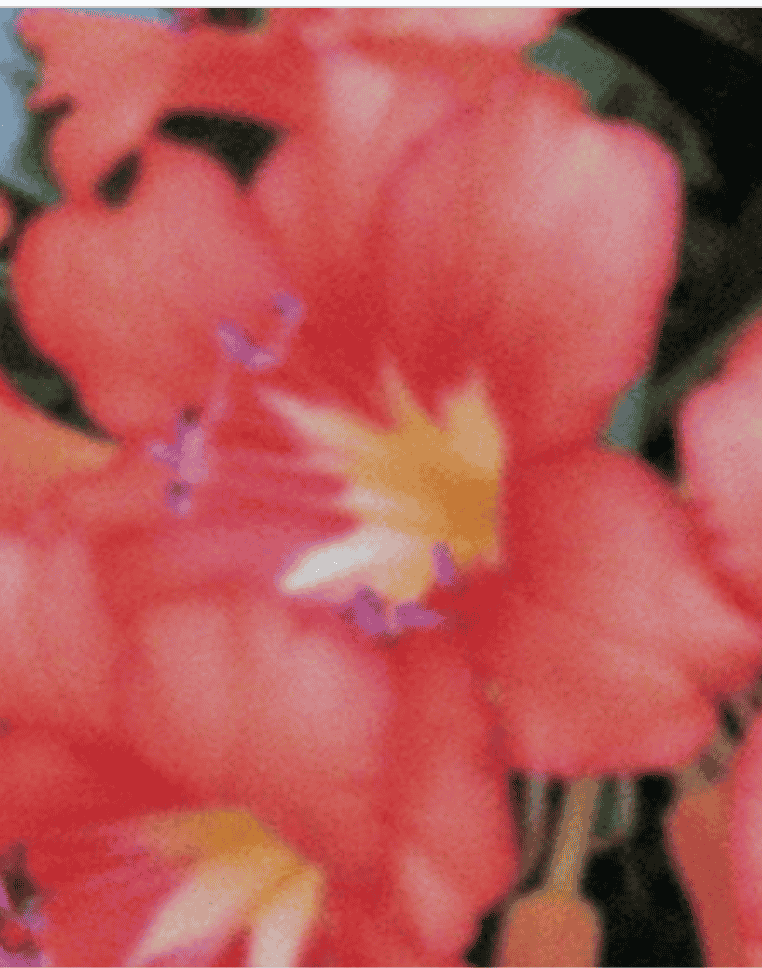}
        \caption{SDS Flower (Zoomed)}
    \end{subfigure}
    \begin{subfigure}[b]{0.24\textwidth}
        \includegraphics[width=\linewidth]{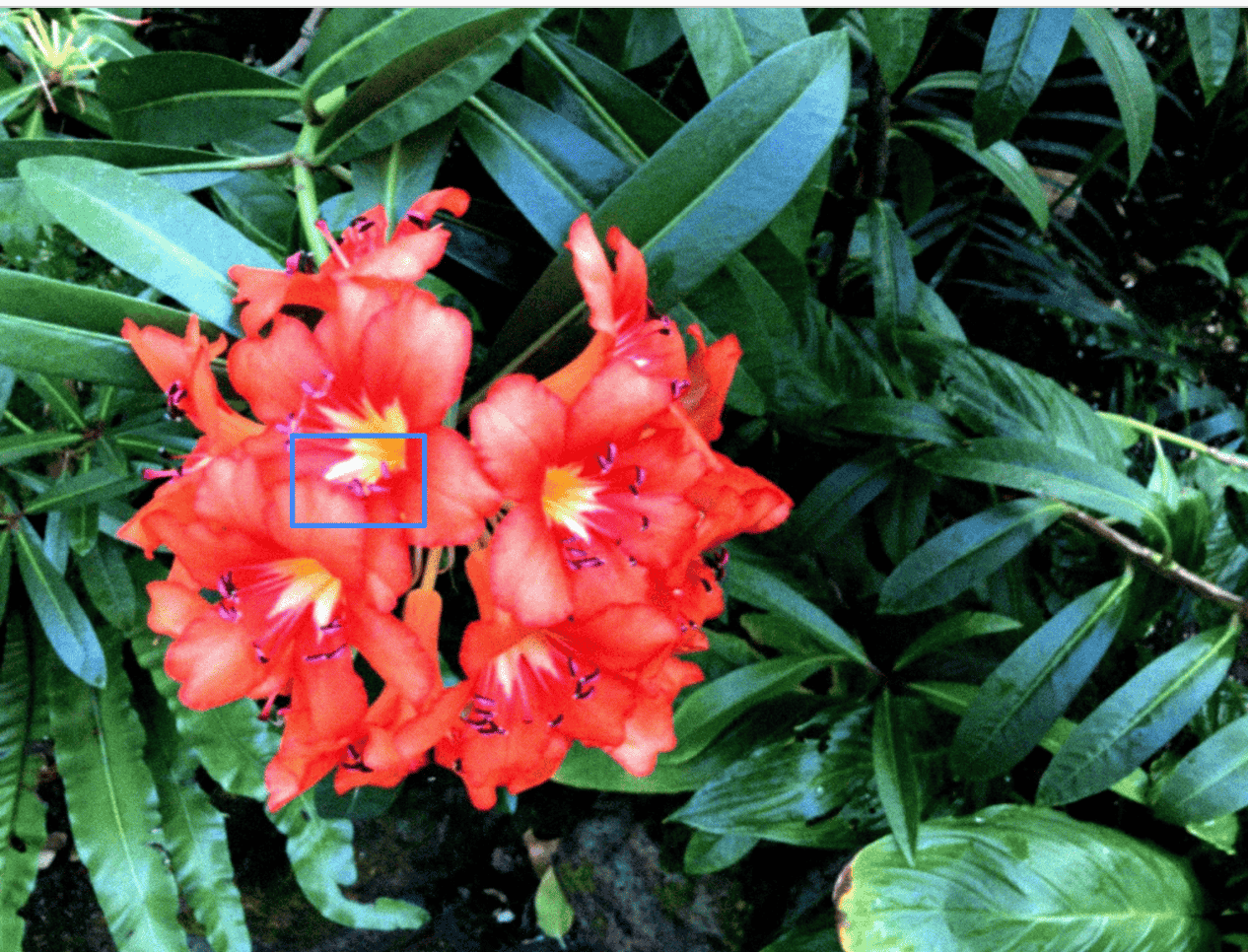}
        \caption{VSD Flower Results}
    \end{subfigure}
    \begin{subfigure}[b]{0.24\textwidth}
        \includegraphics[width=\linewidth]{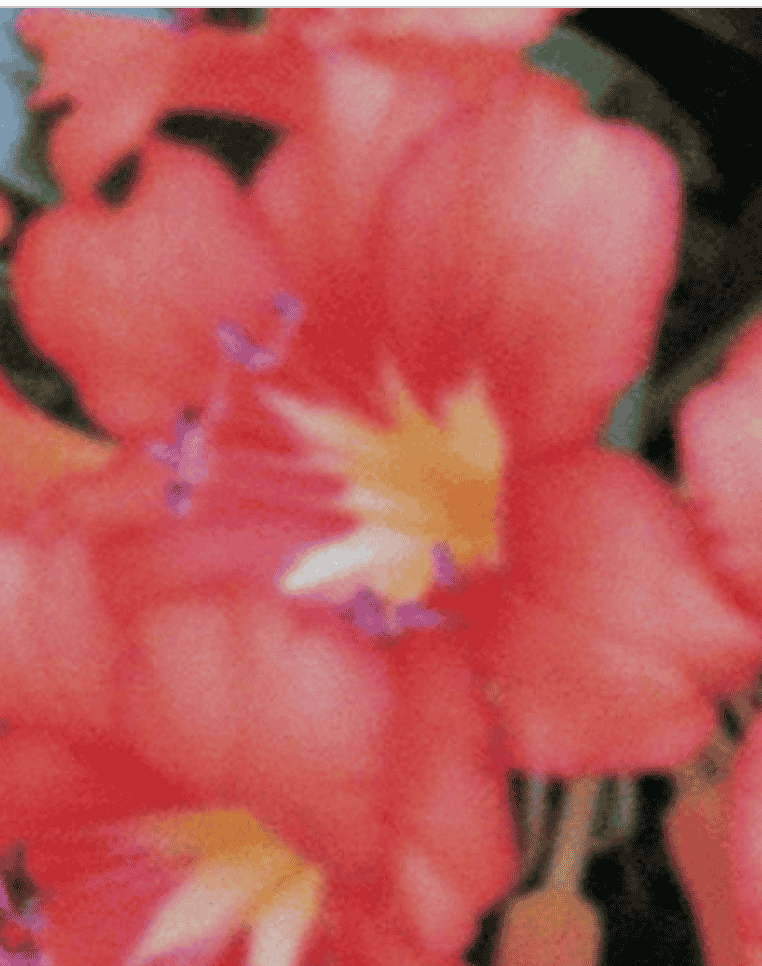}
        \caption{VSD Flower (Zoomed)}
    \end{subfigure}
    \caption{Comparison of SDS and VSD on Flower Images}
    \label{fig:flower-comparison}
\end{figure*}

\begin{figure*}[t]
    \centering
    \begin{subfigure}[b]{0.24\textwidth}
        \includegraphics[width=\linewidth]{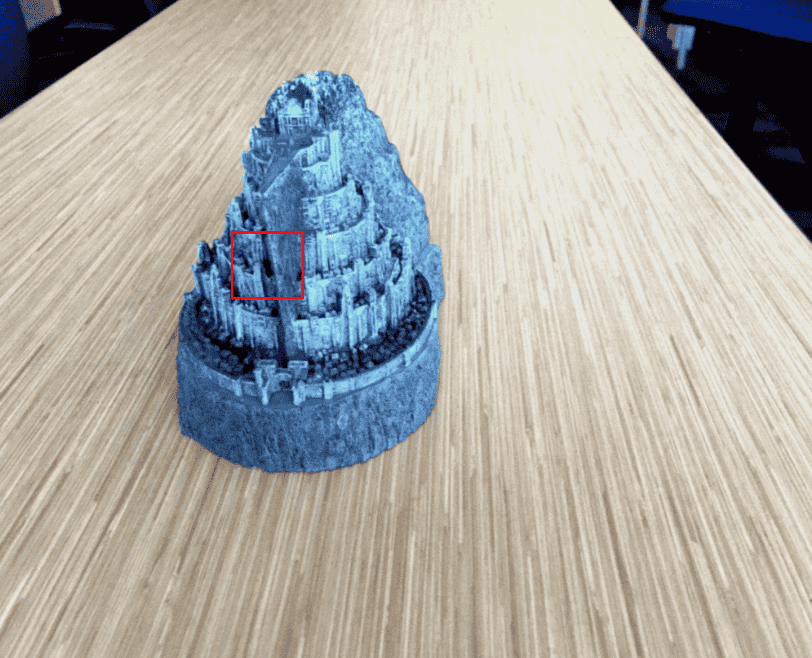}
        \caption{VSD Tower View 1}
    \end{subfigure}
    \begin{subfigure}[b]{0.24\textwidth}
        \includegraphics[width=\linewidth]{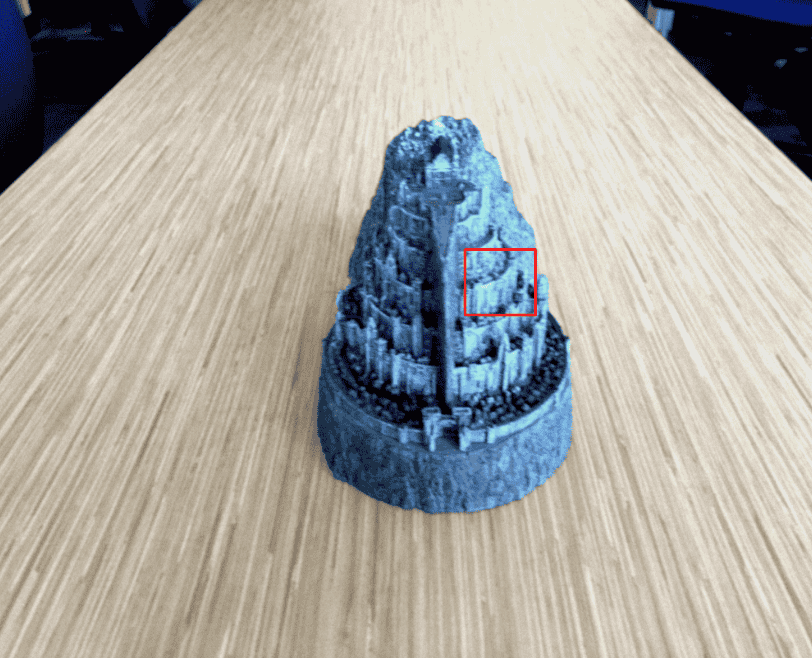}
        \caption{VSD Tower View 2}
    \end{subfigure}
    \begin{subfigure}[b]{0.24\textwidth}
        \includegraphics[width=\linewidth]{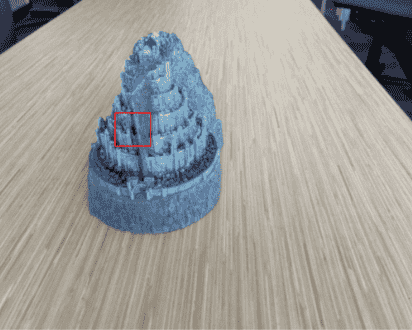}
        \caption{RSD Tower View 1}
    \end{subfigure}
    \begin{subfigure}[b]{0.24\textwidth}
        \includegraphics[width=\linewidth]{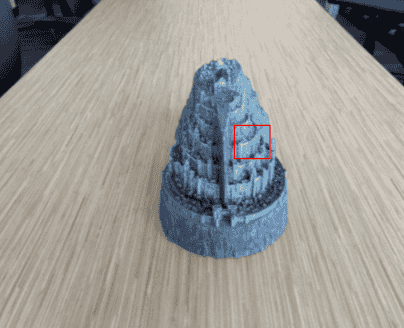}
        \caption{RSD Tower View 2}
    \end{subfigure}
    
    \begin{subfigure}[b]{0.24\textwidth}
        \includegraphics[width=\linewidth]{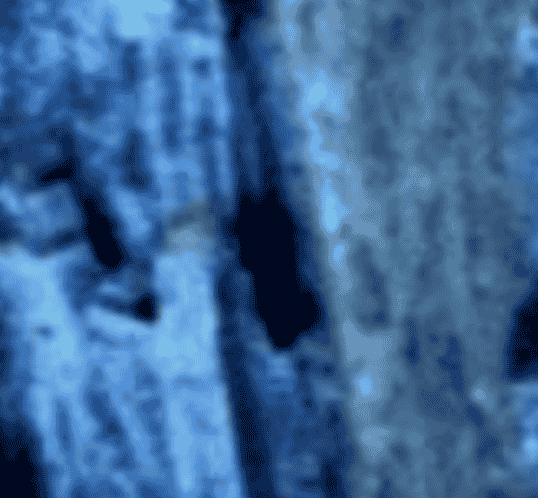}
        \caption{VSD Tower View 1 (Zoomed)}
    \end{subfigure}
    \begin{subfigure}[b]{0.24\textwidth}
        \includegraphics[width=\linewidth]{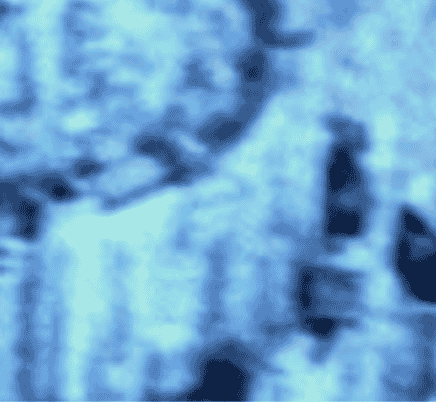}
        \caption{VSD Tower View 2 (Zoomed)}
    \end{subfigure}
    \begin{subfigure}[b]{0.24\textwidth}
        \includegraphics[width=\linewidth]{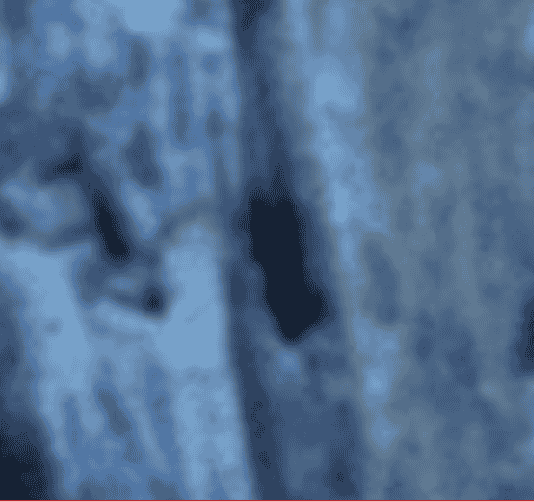}
        \caption{RSD Tower View 1 (Zoomed)}
    \end{subfigure}
    \begin{subfigure}[b]{0.24\textwidth}
        \includegraphics[width=\linewidth]{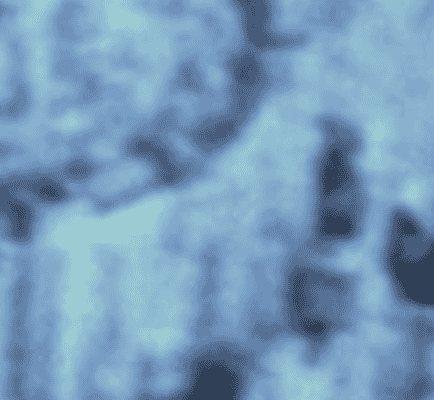}
        \caption{RSD Tower View 2 (Zoomed)}
    \end{subfigure}
    \caption{Comparison of VSD and RSD on Tower Images}
    \label{fig:tower-comparison}
\end{figure*}

First, when comparing the museum artifact image generated with VSD and RSD, there is a clear difference in detail and color saturation. The RSD-generated image has lower color saturation than the VSD-generated image, and furthermore, the VSD-generated image has higher resolution and detail. 

Next, comparing the flower images generated with VSD and SDS, there are noticable differences in image distortion and contrast. The SDS-generated image is has less color contrast than the VSD-generated image. Moreover, the VSD image is less distorted and when zoomed in, shows greater detail than the SDS image. 

A similar theme exists with figures 12-19. Two VSD-generated images are shown along with zoomed in counterparts, and two RSD-generated photos are shown with zoomed-in counterparts. Comparing VSD 1 (Fig. 14) and RSD 1 (Fig. 16), the VSD-generated image has better resolution and more color saturation. Similarly, when comparing VSD 2 (Fig. 15) with RSD 2 (Fig. 19), there is also more detail and better color saturation in the VSD-generated image than there is in the RSD-generated image.

\subsection{Statistical Analysis}
We conducted statistical tests to validate the effectiveness of our proposed method, demonstrating significant improvements in image resolution, detail clarity, and consistency over existing techniques. The evaluation of the NeRFs generated by our model was based on three standard metrics: LPIPS (Learned Perceptual Image Patch Similarity), NIQE (Natural Image Quality Evaluator), and PSNR (Peak Signal-to-Noise Ratio).

\subsection{Comparison with Existing Methods}

We compared our method's performance with existing techniques using LPIPS, NIQE, and PSNR scores. Each of these tests had 10000 initial NeRF render steps with InstantNGP, then 4 rounds of 2000 Super Resolution steps with the mentioned techniques, then we kept 4000 steps of Iterative 3D Synchronization, and arrived at the benchmarks. One interesting finding was that doing spaced training with LoRA, that is training the LoRA every 3 steps instead of 1, it led to slightly better results. The results are summarized in the table below:

\begin{table}[ht]
\centering
\caption{Comparison of Proposed Method with Existing Methods}
\begin{tabular}{lccc}
\toprule
Method  & LPIPS & NIQE   & PSNR   \\
\midrule
No Changes (Plain RSD) & \textbf{0.1496} & 4.983 & 3.983 \\
With SDS & 0.1587 & 5.667 & 3.529 \\
With VSD + LoRA Spaced  & 0.1523 & \textbf{4.457} & \textbf{4.026} \\
With VSD + LoRA & 0.1550 & 4.612 & 3.998 \\
\bottomrule
\end{tabular}
\end{table}

As we can see, our methods outperform on NIQE and PSNR, showing our images our higher quality and more natural. Where we do lack a little bit is in LPIPS, and this is explained further in our Limitations section, but we believe it has to do with VSD giving us higher contrast images, which look much different from the ground truth, an issue seen in other projects as well with VSD, such as ProlificDreamer.

\subsection{Limitations}

While our approach significantly enhances the quality and consistency of Neural Radiance Fields (NeRF) using Variational Score Distillation (VSD) and Low-rank Adaptation (LoRA), it has some limitations. 

Specifically, the Learned Perceptual Image Patch Similarity (LPIPS) score, though improved from the baseline, remains lower than that of Renoised Score Distillation (RSD). This indicates that while our model excels in high resolution and view consistency, it may not match RSD in perceptual fidelity. The VSD's emphasis on higher contrast can lead to overemphasis on certain features, potentially compromising perceptual coherence.

In summary, our methodology achieves notable improvements in resolution and consistency but could benefit from further refinement to enhance perceptual quality. Future work should address these limitations to better balance resolution, consistency, and perceptual accuracy.

To add, due to the backpropagation of the LoRA parameters, our method is 15 to 20\% slower than RSD, but this is negligible due to advancements in optimization with backpropgation.

\section{Conclusion}
In this paper, we introduced a new method for improving Neural Radiance Fields (NeRF) with Super Resolution by using Variational Score Distillation (VSD) and I3DS. Our approach significantly enhances the resolution and detail of images while maintaining consistency in the generated outputs.

Our experimental results confirm that our method outperforms traditional techniques. By integrating advanced technologies such as VSD + LoRA and I3DS, we achieved higher quality images that are closer to real-life visuals. The metrics LPIPS, NIQE, and PSNR showed clear improvements in image quality and consistency compared to existing methods.

The practical implications of our findings are vast. They can benefit various applications in 3D modeling, virtual reality, and computer graphics by providing more accurate and realistic images. This advancement could lead to more immersive experiences in gaming and simulations, as well as improved accuracy in professional fields like medical imaging and architectural visualization.

Future research could focus on further refining these techniques and exploring their application in different contexts or with different types of data. Additional work on reducing computational demands and increasing processing speed could make these improvements more accessible for real-time applications.

In conclusion, our method sets a new standard for NeRF editing, promising more realistic and consistent 3D image generation. 

{\small
\bibliographystyle{ieee_fullname}
\bibliography{egbib}
}

\end{document}